\documentclass[10pt,twocolumn,letterpaper]{article}

\usepackage{iccv}              
\usepackage{algorithm}
\usepackage{algpseudocode}
\usepackage{xcolor}

\definecolor{iccvblue}{rgb}{0.21,0.49,0.74}
\definecolor{codehighlight}{rgb}{0.16,0.61,0.36}

\usepackage[pagebackref,breaklinks,colorlinks,allcolors=iccvblue]{hyperref}
\usepackage{multirow}
\usepackage{amssymb}
\usepackage{graphicx}
\usepackage{subcaption}

\def\paperID{14369} 
\def\confName{ICCV}
\def\confYear{2025}

\title{Active Learning Inspired ControlNet Guidance\\for Augmenting Semantic Segmentation Datasets}

\author{Hannah Kniesel\\
Visual Computing Group\\
Ulm University\\
{\tt\small hannah.kniesel@uni-ulm.de}
\and
Pedro Hermosilla\\
Computer Vision Lab\\
TU Vienna\\
{\tt\small phermosilla@cvl.tuwien.ac.at}
\and
Timo Ropinski\\
Visual Computing Group\\
Ulm University\\
{\tt\small timo.ropinski@uni-ulm.de}
}

\begin{document}
\maketitle
\begin{abstract}
Recent advances in conditional image generation from diffusion models have shown great potential in achieving impressive image quality while preserving the constraints introduced by the user. In particular, ControlNet enables precise alignment between ground truth segmentation masks and the generated image content, allowing the enhancement of training datasets in segmentation tasks. This raises a key question: Can ControlNet additionally be guided to generate the most informative synthetic samples for a specific task? Inspired by active learning, where the most informative real-world samples are selected based on sample difficulty or model uncertainty, we propose the first approach to integrate active learning-based selection metrics into the backward diffusion process for sample generation. Specifically, we explore uncertainty, query by committee, and expected model change, which are commonly used in active learning, and demonstrate their application for guiding the sample generation process through gradient approximation. Our method is training-free, modifying only the backward diffusion process, allowing it to be used on any pretrained ControlNet. Using this process, we show that segmentation models trained with guided synthetic data outperform those trained on non-guided synthetic data. Our work underscores the need for advanced control mechanisms for diffusion-based models, which are not only aligned with image content but additionally downstream task performance, highlighting the true potential of synthetic data generation. 
\end{abstract}   
\section{Introduction}
\label{sec:intro}

The rapid advancements in generative models~\cite{rombach2022high,podell2023sdxl,radford2021learning,ramesh2021zero,saharia2022photorealistic} have transformed the landscape of data generation, presenting unprecedented opportunities for enhancing training with additional data or even training on synthetic images only~\cite{azizi2023synthetic,sariyildiz2023fake,zhou2023using,wang2023better,schnellscribblegen, ye2025seggen, jia2025dginstyle, kupyn2024dataset,hammoud2024synthclip}. Today, these models can not only be used to generate new data samples but also their corresponding labels~\cite{ho2022classifier,dhariwal2021diffusion,ye2025seggen,jia2025dginstyle,kupyn2024dataset,wu2023diffumask,nguyen2024dataset,hammoud2024synthclip}, making them especially valuable for tasks where manual annotation is resource-intensive, such as semantic segmentation. Especially, ControlNet~\cite{zhang2023adding} has great potential in this domain, and many techniques for enhancing segmentation datasets build upon it~\cite{schnellscribblegen, ye2025seggen, jia2025dginstyle, kupyn2024dataset}. By being able to generate synthetic training data with labels, manual labeling -- traditionally a major bottleneck in computer vision research -- is reduced significantly.

Besides the possibility to enhance training datasets without additional labeling cost, generative models enable to guide image content based on predefined principles. This means that unlike real-world images, where we are limited to selecting training samples from a fixed set, this allows us to guide the image generation process toward desired samples when working with synthetic images. In this context, a fundamental question arises: Can we guide the generation process to produce samples that are inherently more valuable for training? 

Targeted image generation has been explored in prior work to improve alignment with text prompts or enhance image quality~\cite{kim2024datadream,rombach2022high,dhariwal2021diffusion,dhariwal2021diffusion,ho2022classifier,chen2024training,singh2023divide,feng2022training,chefer2023attend,dai2025advdiff}. However, to the best of our knowledge, none of these approaches directly aim to guide sample generation toward creating more useful samples for downstream model performance.

In contrast, various methods~\cite{shum2025predictive,zhu2024generative,fan2024divergen,lampis2023bridging,ye2020synthetic,li2023data,alaa2022faithful} have used subsampling or data selection to improve downstream training efficiency and performance. Subsampling techniques have been studied for both real-world and synthetic data based on a range of criteria. More specifically, subsampling a large data pool to extract the most valuable samples directly intersects with active learning~\cite{settles.tr09}, a strategy that prioritizes the most informative samples for annotation, improving model performance while minimizing labeling effort. Active learning has proven effective for real-world images as well as synthetic data by reducing annotation costs and computational overhead as well as improving model performance~\cite{mahapatra2018efficient,zhu2024generative,sener2017active,yang2024plug,garcia2023ten,xie2020deal,ren2021survey}.

Building on these approaches, we propose a new research direction: As generative models improve, the need for additional guidance toward realism diminishes. Instead, we focus on guiding the generation process toward the most informative samples for downstream task performance. This shifts the focus from post-generation filtering to proactive, targeted generation of valuable synthetic data.

Within this paper, we investigate whether active learning principles -- such as sample difficulty and model uncertainty -- can be leveraged to guide the generation of synthetic datasets for semantic segmentation. Specifically, we examine three common active learning query strategies: First, we assess \textit{uncertainty}~\cite{lewis1995sequential} by using entropy~\cite{shannon1948mathematical}. Second, we approximate \textit{query by committee}~\cite{seung1992query} by leveraging Bayesian approximation with Monte Carlo dropout layers~\cite{gal2016dropout}. Lastly, we apply \textit{expected model change}~\cite{settles2007multiple} by computing the cross-entropy loss. 
Our findings demonstrate that incorporating active learning strategies into the generation process can lead to more informative samples, ultimately improving the downstream performance of segmentation models.

Our approach is innovative in that it remains training-free and directly guides the diffusion process of generative models during inference, targeting data generation in a more strategic manner. Specifically, we contribute the following:

\begin{itemize}
    \item We introduce active learning inspired ControlNet guidance of the backward diffusion process, to obtain synthetic images which are more valuable for downstream task performance.

    \item We reveal that entropy is able to outperform other active learning inspired metrics for the proposed ControlNet guidance.  
    
    \item Our evaluation shows that training on guided synthetic data is able to outperform training on non-guided synthetic data.
\end{itemize}

In summary, this work presents a novel approach for generating synthetic data optimized for semantic segmentation, leveraging the unique properties of synthetic data and generative models to strategically enhance dataset quality for downstream model training.

\section{Related Work}
\label{sec:relatedwork}
Within this section we highlight relevant works in the field of diffusion models, diffusion models for segmentation, guidance of diffusion models and active learning, as these domains lay the foundation for our ControlNet Guidance. 

\subsection{Conditional Image Generation}

\paragraph{Diffusion Models.}
Diffusion models, such as \textit{Stable Diffusion}~\cite{rombach2022high}, \textit{DALL-E}~\cite{ramesh2021zero}, and \textit{Imagen}~\cite{saharia2022photorealistic}, have gained attention for generating high-quality samples by learning to reverse the process of adding noise to images. These models also enable conditional generation, often through text prompts. Latent diffusion models~\cite{rombach2022high} further reduce computational complexity by operating in latent space using a VAE. Diffusion models are applied in synthetic image generation, dataset augmentation, and improving adversarial robustness. For example, Azizi et al.\cite{azizi2023synthetic} used Imagen~\cite{saharia2022photorealistic} for augmenting datasets, Zhou et al.\cite{zhou2023using} showed how Stable Diffusion-based augmentation boosts classification accuracy and ~\citet{sariyildiz2023fake} demonstrated that Stable Diffusion could produce synthetic training data with generalization capabilities comparable to models trained on real data. Recent studies also highlight their potential in adversarial defense~\cite{wang2023better}.

\paragraph{Diffusion Models for Segmentation.}\label{subsec:segmentation_diffusion}
Diffusion models have shown promise in segmentation tasks. For instance, \textit{DiffuMask}\cite{wu2023diffumask} uses text-guided cross-attention for pixel-wise mask generation, achieving state-of-the-art results in open-vocabulary segmentation. Similarly, \textit{Dataset Diffusion} improves segmentation quality by leveraging self- and cross-attention maps~\cite{nguyen2024dataset}.

\textit{ControlNet}~\cite{zhang2023adding} enables conditioning Stable Diffusion on structured data like edges and depth, providing more control over the generation process. This has been applied to data augmentation for segmentation, as in \textit{SegGen}~\cite{ye2025seggen}, which generates images conditioned on segmentation masks, and \textit{ScribbleGen}~\cite{schnellscribblegen}, which improves segmentation with scribble-based guidance. Similarily, \textit{DGInStyle}~\cite{jia2025dginstyle} enhances data diversity for domain-generalizable segmentation leveraging ControlNet.

Despite the additional conditioning capabilities of ControlNet,~\citet{kupyn2024dataset} found that diffusion models struggle to generate complex scenes with multiple foreground and background objects. To address this limitation and ensure precise alignment between generated images and their corresponding ground truth masks, they propose a structured approach: first, annotations are decomposed into per-object binary masks. Then, an inpainting ControlNet, conditioned on edge and (predicted) depth maps, is used to redraw each instance individually. Finally, the instances are recombined into a coherent scene using depth-based alpha blending. This method significantly improves model training across various tasks, including object detection, semantic segmentation, and instance segmentation, yielding strong results. 

While these approaches show promise, they don’t explicitly optimize segmentation model performance during data generation, which motivates our approach to directly guide the generation process with ControlNet Guidance for better segmentation outcomes.

\paragraph{Diffusion Model Guidance.}
Diffusion model guidance has been studied in a large variety of tasks. One of the most prominent guidance methods for diffusion models is \textit{classifier-guided diffusion}~\cite{dhariwal2021diffusion}. In this approach, an auxiliary classifier is trained to assess whether an intermediate diffusion step aligns with the desired attributes, allowing it to guide the generation process for more directed outputs. However, classifier guidance requires the training of an additional classifier, adding computational overhead. Instead, Ho et al.~\cite{ho2022classifier} proposed \textit{classifier-free guidance}: This approach eliminates the need for the extra classifier network, by directing the image generation process using a conditional input, such as a text prompt, thereby improving the relevance and coherence of generated images.

In another line of work, guidance has been shown crucial in achieving better alignment between image generation and text prompts. \textit{Training Free Layout Guidance}~\cite{chen2024training} addresses layout issues by using cross-attention and bounding box prompts during inference to improve layout alignment. Similarly,~\citet{singh2023divide} detect and prioritize misaligned parts of generated images within a modified reverse diffusion process, they are able to text-to-image alignment.~\citet{feng2022training} improve compositional synthesis by using scene graphs to capture spatial and relational aspects in prompts. Additionally, \citet{chefer2023attend} introduce \textit{Generative Semantic Nursing (GSN)} to amplify activations associated with prompt tokens, improving image synthesis accuracy.

Further, guided diffusion is also employed in generating adversarial images for classification tasks, producing samples with specific properties. Methods like \textit{AdvDiff}\cite{dai2025advdiff} and\citet{chen2024diffusion} generate imperceptible, transferable adversarial attacks using guidance. \textit{AdvDiffuser}~\cite{chen2023advdiffuser} synthesizes natural adversarial examples, while \citet{chen2024content} propose a content-based attack for precise control over adversarial generation. Guided diffusion is also employed for adversarial purification, helping remove perturbations from images affected by attacks. For instance, \citet{lin2024robust} propose a purification method using diffusion models to counter adversarial effects, improving model robustness.

These works underscore the potential of guided diffusion to refine prompt alignment, produce controlled adversarial samples, and purify images impacted by adversarial perturbations. Motivated by this potential, within this paper we introduce the first guidance, which directly targets downstream performance of semantic segmentation models, trained on synthetic data.

\subsection{Active Learning}
Active learning is a machine learning technique designed to enhance model performance while reducing labeling costs by strategically selecting the most informative samples from an unlabeled dataset. This technique focuses on samples that provide the greatest improvement to the model, making it especially valuable when labeling is costly or time-consuming. Active learning has been successfully applied across various domains, from traditional machine learning~\cite{settles2007multiple,tong2001support,settles2008analysis} to large language models~\cite{astorga2025active,carvalho2025deep,wang2024method} and deep learning for computer vision~\cite{yang2017suggestive,xie2022towards,yuan2021multiple}.

Key query strategies in active learning include \textit{uncertainty}\cite{lewis1995sequential}, \textit{query by committee}\cite{seung1992query}, and \textit{expected model change}~\cite{settles2007multiple}.
In the uncertainty-based strategy, entropy~\cite{shannon1948mathematical} is often used to measure the model’s uncertainty over predicted classes. A high entropy indicates that the model is unsure about its prediction, making the sample more informative.
In the query by committee approach, multiple models make predictions on the same input, and if there is significant disagreement among the models, the sample is considered informative and selected for further annotation and model training. However, this approach can be computationally expensive due to the need for a large committee. To mitigate this, Monte Carlo dropout~\cite{gal2016dropout} can be used as a Bayesian approximation, effectively simulating the behavior of a committee with a smaller computational overhead.
The expected model change strategy focuses on selecting samples that are expected to cause the largest change in the model’s parameters. This is typically done by selecting those with the highest computed error, which indicates the greatest potential for improvement.

\section{Background}
In the following, we describe the basic principles of diffusion models and ControlNet, as these methods form the foundation of our proposed active learning-inspired ControlNet guidance.  

\subsection{Diffusion Models}  
Diffusion models are a class of generative models that produce samples by gradually transforming Gaussian noise into a structured image through an iterative denoising process. In the standard latent diffusion framework, a model generates samples by progressively refining a noise sample \( \mathbf{x}_T \sim \mathcal{N}(0,1) \) over \( T \) denoising steps until it produces a final, clean image \( \mathbf{x}_0 \).  

At each timestep \( t \in [0,T] \), the latent representation \( \mathbf{x}_t \) consists of a mixture of an underlying clean image estimate \( \hat{\mathbf{x}}_0 \) and a noise component \( \boldsymbol{\epsilon}_t \). A trained denoising model, typically a U-Net, predicts the noise \( \boldsymbol{\epsilon}_t \), which can then be subtracted from \( \mathbf{x}_t \) to estimate the original clean image. The denoising update step in a standard diffusion model is given by:  

\begin{equation}
\mathbf{x}_{t-1} = \frac{1}{\sqrt{\alpha_t}} \left( \mathbf{x}_t - \boldsymbol{\epsilon}_t \right),
\end{equation}

where \( \alpha_t \) represents the noise schedule, which controls the level of noise at each timestep. The full denoising process is performed sequentially over multiple steps, gradually removing noise and refining the image structure.  

While diffusion models have demonstrated strong generative capabilities, their standard formulation lacks direct control over the generated content, which can be problematic for tasks requiring precise structure or guidance. This limitation is addressed by ControlNet, which introduces additional conditioning mechanisms to steer the generation process.  

\subsection{ControlNet}  
ControlNet~\cite{zhang2023adding} extends diffusion models by enabling explicit structural conditioning through auxiliary inputs such as edge maps, segmentation masks, depth maps, or other image features.
During training, ControlNet learns a mapping from these control inputs to the desired image output while leveraging the pre-trained weights of a standard diffusion model, such as Stable Diffusion~\cite{rombach2022high}. By incorporating structural priors, it enhances the quality and alignment of generated images, making it particularly valuable for data augmentation in segmentation tasks.

Despite its effectiveness,~\citet{kupyn2024dataset} found that ControlNet struggles with generating complex scenes containing multiple foreground and background objects. To address this, they proposed training and applying ControlNet in an inpainting fashion, redrawing individual instances separately to improve generation fidelity.

\section{Method}
\label{sec:method}
The core of our proposed method integrates an active learning-inspired guidance mechanism into ControlNet, refining its image generation to produce the most informative samples for downstream model training. By leveraging a pretrained ControlNet, we ensure high-fidelity image generation while prioritizing sample informativeness. Our training-free approach seamlessly incorporates guidance into the backward diffusion process, modifying only three lines of code—highlighting its simplicity and elegance.

\subsection{Overview}
\begin{figure}[t]
  \centering
  \includegraphics[width=1.0\linewidth]{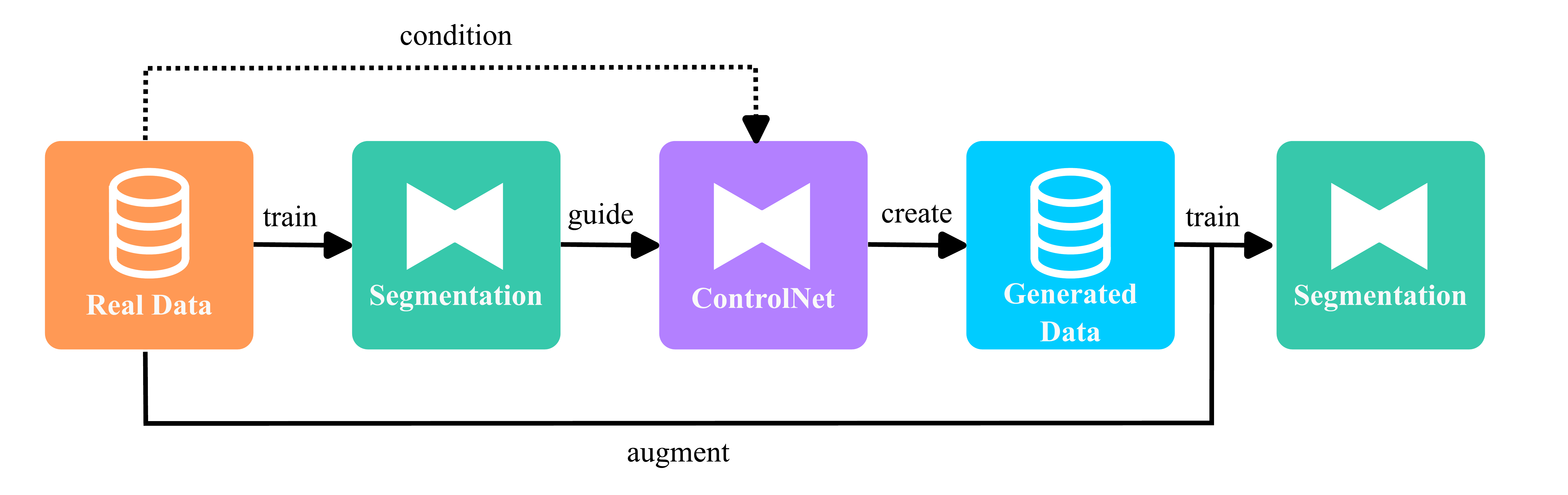}

   \caption{Our proposed iterative data generation and model refinement pipeline introducing active learning inspired ControlNet guidance: A segmentation model trained on real data guides ControlNet to generate informative, real-data-aligned samples, which are then added to the training set for model retraining. This active learning-inspired process refines the model through data generation.}
   \label{fig:overview}
\end{figure}

\autoref{fig:overview} illustrates our approach, which follows an iterative process of model training, guided data generation, and retraining, akin to nowadays active learning frameworks. We begin by training a segmentation model on the initial dataset. We then adapt the ControlNet introduced in~\cite{kupyn2024dataset} with our proposed active learning inspired ControlNet guidance. Thus we incorporate the segmentation model's predictions to guide the generation process, targeting samples that are expected to be most beneficial for improving model performance. This approach aligns with active learning principles, where the most informative or uncertain samples are selected for retraining~\cite{settles.tr09}. Finally, the generated data is used to augment the original dataset, and the segmentation model is retrained, progressively improving its accuracy and robustness.

\subsection{ControlNet Guidance}
Similar to classifier-free guidance~\cite{ho2022classifier}, we modify the latent representation $x_t$ at each diffusion denoising step $t\in[0,T]$ to drive it towards a more desirable representation $\hat{x}_t$. This approach enforces specific characteristics in the generated data. In the following we will refer to this technique as \textit{latent guidance}. In this paper, we present, for the first time, how to apply latent guidance to ensure that ControlNet-generated images conform to constraints that improve downstream task performance.

\paragraph{Guidance via Gradient Optimization.}
During latent guidance, we modify the current latent representation \( \mathbf{x}_t \) to steer the final image \( \mathbf{x}_0 \) towards a predefined objective. This is done by introducing an additional gradient-based update:

\[
\mathbf{\hat{x}}_{t} = \mathbf{x}_t - \eta_t \nabla_{\mathbf{x}_t} \mathcal{L}(\hat{\mathbf{x}}_0),
\]

where \( \mathcal{L}(\hat{\mathbf{x}}_0) \) represents a task-specific loss function computed on the estimated clean image \( \hat{\mathbf{x}}_0 \). The step size \( \eta_t \) controls the strength of the guidance at each denoising diffusion step $t$. We will further refer to this as guidance strength.

\paragraph{Guidance Loss.}
To guide the generative process, we incorporate feedback from a pre-trained segmentation model, $g_{\phi}$, which is trained on real data and provides predictions for synthetic samples. This model allows us to evaluate synthetic data by measuring sample difficulty as well as uncertainty inspired by active learning practice. 

To do so, we investigate three commonly used active learning query strategies for guiding the image generation. We investigate query by committee, expected model change and uncertainty~\cite{settles.tr09}. 

For computing the \textit{query by committee}, we use Monte Carlo dropout as Bayesian approximation, simulating the behavior of a committee. We compute the disagreement of the committee by the variance of model predictions across multiple stochastic forward passes with activated dropout layers. This variance is given by:

\[
\text{MCD} = \frac{1}{N} \sum_{n=1}^{N} \left( g_{\phi}^{(n)}(\hat{\mathbf{x}}_0) - \bar{g}_{\phi}(\hat{\mathbf{x}}_0) \right)^2
\]

where \( g_{\phi}^{(n)}(\hat{\mathbf{x}}_0) \) is the prediction in the \( n \)-th forward pass, and \( \bar{g}_{\phi}(\hat{\mathbf{x}}_0) \) is the mean prediction over \( N \) passes. A high variance indicates high disagreement of the committee usually highlighting more informative samples.

To guide based on the \textit{expected model change} we employ the Cross-Entropy (CE) loss, measuring the discrepancy between the model’s predicted segmentation mask \( g_{\phi}(\hat{\mathbf{x}}_0) \) for the input image \( \hat{\mathbf{x}}_0 \) and the ground truth mask \( \mathbf{y} \). Samples which cause a high CE loss are expected to cause the largest change in the model’s parameters. It is defined as:

\[
\text{CE} = - \sum_{c} \mathbf{y}_c \log g_{\phi}(\hat{\mathbf{x}}_0)_c
\]

where \( c \) represents the class index. By leveraging this metric, we aim to maximize sample difficulty by generating images where the model prediction deviates significantly from the ground truth.

Finally, for guiding the image generation using \textit{uncertainty}, we apply entropy, which quantifies the uncertainty over predicted class probabilities. Thus, we guide the generation towards high entropy, making it difficult for the model to confidently assign a single class label. The entropy is defined as:

\[
\text{Entropy} = - \sum_{c} g_{\phi}(\hat{\mathbf{x}}_0)_c \log g_{\phi}(\hat{\mathbf{x}}_0)_c
\]

where high entropy corresponds to greater uncertainty in the model’s predictions.

In the following we will refer to the different guidance loss functions as \textit{MCD}, \textit{CE} and \textit{Entropy} respectively.

\paragraph{Gradient Approximation for Guidance.}
The introduced guidance functions all depend on the segmentation model predictions on the clean image $x_0$. This clean image is not available for each denoising step. Instead, we estimate the clean image $\hat{x_0}$ within each denoising step based on the current latent $x_t$, following~\citet{song2020denoising}. 

\[
\hat{\mathbf{x}}_{0} \approx \frac{1}{\sqrt{\alpha_t}} \left( \mathbf{x}_t - \boldsymbol{\epsilon}_t \right),
\]

\begin{figure}[h]
  \centering
  \includegraphics[width=1.0\linewidth]{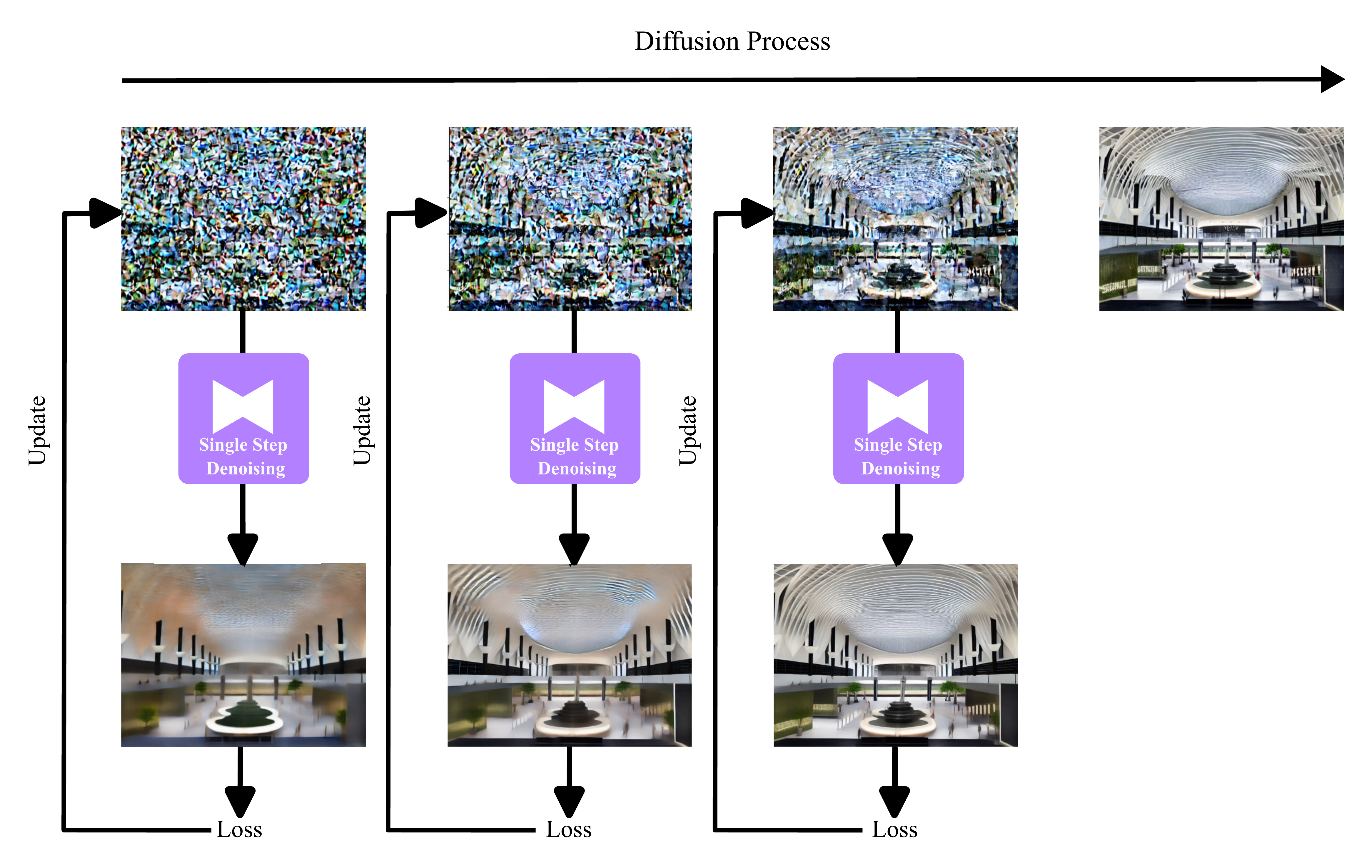}

   \caption{Visualization of latent guidance with single step denoising. For each denoising step, we approximate the clean image $\hat{x_0}$ by single step denoising, such that we are able to apply the loss function to update the current latent $x_t$.}
   \label{fig:guidance}
\end{figure}

We refer to this as \textit{single step denoising} shown in~\autoref{fig:guidance}.

\paragraph{Algorithm.}
Our latent guidance approach introduces minimal computational overhead to the standard backward diffusion process. In the following latent guidance pseudocode, we highlight the three additional computations required to apply our active learning-inspired ControlNet guidance in green. These three steps are the only modifications needed compared to the standard backward diffusion process.

\begin{algorithm}[H]
\caption{Latent Guidance in Diffusion Models}
\begin{algorithmic}
\State \textbf{Input:} Trained UNet model, initial noise sample \( \mathbf{x}_T \sim \mathcal{N}(0,1) \), guidance strength \( \eta \),
\For{\( t = T, \dots, 1 \)}
    \State \( \boldsymbol{\epsilon}_t = \text{UNet}(\mathbf{x}_t, t) \)  \hfill // Predict noise
    \State \textcolor{codehighlight}{\( \hat{\mathbf{x}}_0 = \frac{1}{\sqrt{\alpha_t}} \left( \mathbf{x}_t - \boldsymbol{\epsilon}_t \right) \)}  \hfill // Estimate clean image
    \State \textcolor{codehighlight}{\( \mathbf{g}_t = \nabla_{\mathbf{x}_t} \mathcal{L}(\hat{\mathbf{x}}_0) \)}  \hfill // Compute guidance gradient
    \State \textcolor{codehighlight}{\( \hat{\mathbf{x}}_t = \mathbf{x}_t - \eta_t \cdot \mathbf{g}_t \)}  \hfill // Apply guidance
    \State \( \mathbf{x}_{t-1} = \frac{1}{\sqrt{\alpha_t}} \left( \hat{\mathbf{x}}_t - \boldsymbol{\epsilon}_t \right) \)  \hfill // Perform denoising step
\EndFor
\State \textbf{Output:} Final generated image \( \mathbf{x}_0 \)
\end{algorithmic}
\end{algorithm}

\section{Experiments}
\subsection{Experimental Setup}

\paragraph{Datasets.}  
We evaluate our approach on well-established benchmark datasets for semantic segmentation, specifically COCO-Stuff10k~\cite{caesar2018coco} (COCO10k), which involves predicting 171 different classes, and PascalVOC2012~\cite{everingham2010pascal} (VOC2012), with 20 distinct classes. For the ablation studies, we create a reduced version of COCO10k by randomly subsampling it to 2,500 training images, referred to as COCO2.5k. 

\paragraph{Evaluation Metric.}  
To assess the impact of our guided image generation, we train a downstream segmentation model using the synthetic data as augmentations. We evaluate segmentation performance using two standard metrics: the mean Intersection over Union (mIoU) and the mean accuracy (mAcc). The mIoU measures the average overlap between predicted and ground truth regions, while mAcc calculates the average accuracy across all classes, providing a balanced view of model performance in semantic segmentation. 

\begin{figure*}[t]
  \centering
  \includegraphics[width=1.0\linewidth]{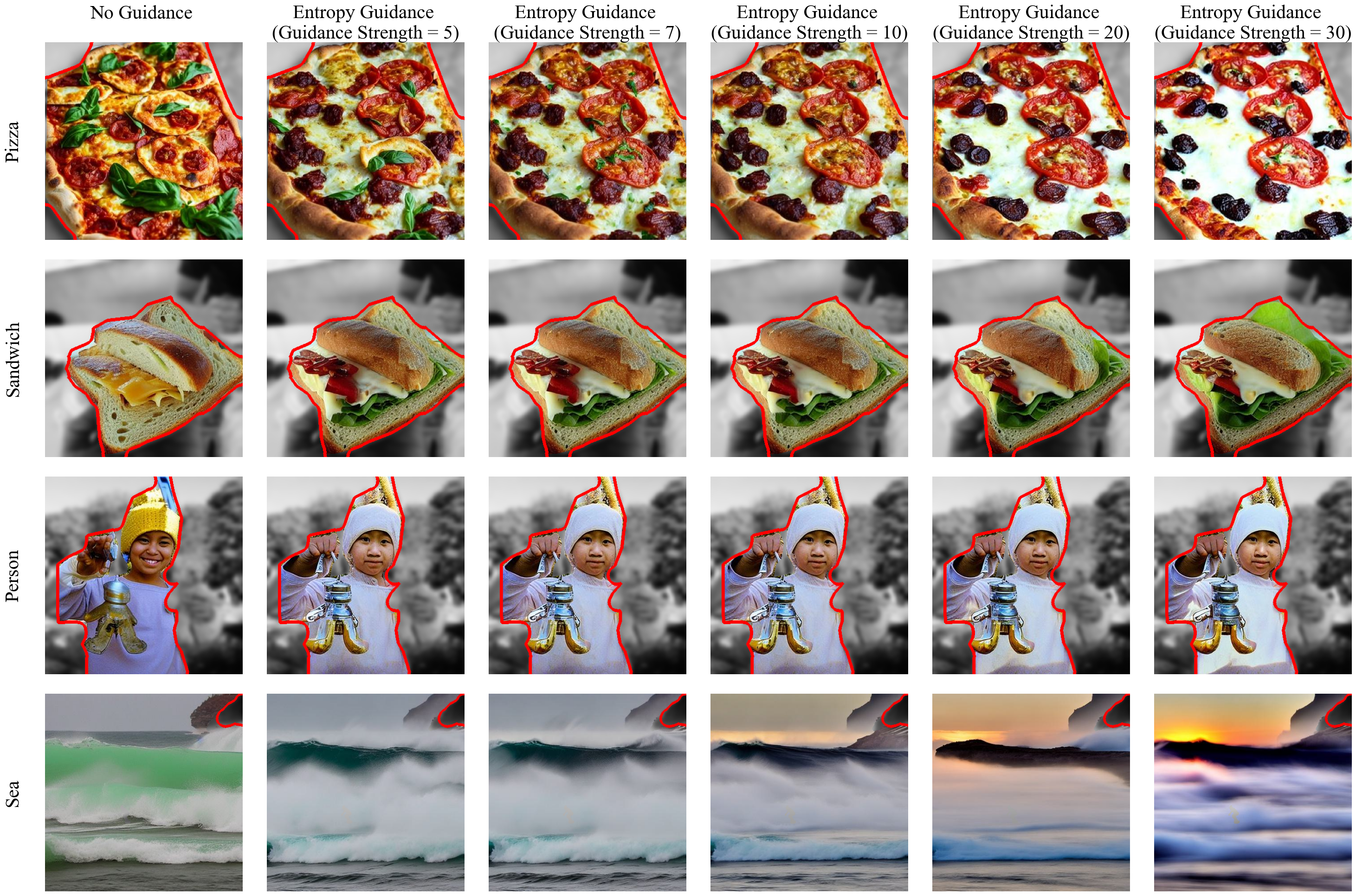}

   \caption{Qualitative results showing the influence of guidance strength. The first column displays the non-guided images, while the subsequent columns show results with our proposed active learning-inspired ControlNet guidance at varying strengths. Red borders highlight the synthetically augmented objects. As the guidance strength increases, more pronounced changes are observed in the images. However, stronger guidance can also lead to unintended alterations in the content, such as the green sandwich bread in the second row for a guidance strength of 30.}
   \label{fig:guidancestrength}
\end{figure*}

\paragraph{Data Generation.}
As stated in~\autoref{subsec:segmentation_diffusion} there are many approaches which have shown the usefulness of ControlNet for synthetically augmenting segmentation datasets. In particular,~\citet{kupyn2024dataset} achieved impressive results by addressing ControlNet’s limitations in generating complex scenes with multiple objects. Hence, in our work, we follow their methodology and extend it to the proposed active learning inspired guidance.  We employ their published ControlNet model with a Stable Diffusion backbone, conditioned on edge maps and (predicted) depth maps.
For data generation, we follow their configuration and use $T=40$ denoising steps. 
We extract per-object binary masks using connected components~\cite{rosenfeld1966sequential} from the annotations.
Unless stated otherwise, we select single instances for augmentation based on their size, prioritizing the largest objects in the image.
Using this procedure, we generate one augmented image per real image, and train the segmentation model with an augmentation probability of $p = 0.5$.
If not stated otherwise, within our experiments, we explored guidance scales $\in[7,10,20,30]$ for all datasets and report the best performing. We show the influence of the guidance strength and ablate different guidance schedulers in the supplementary material (see~\autoref{supp:schedule} and ~\autoref{fig:guidancestrength}). For more details please see our codebase (released upon acceptance).

\paragraph{Segmentation Model.}  
We use SegNext-L~\cite{guo2022segnext} as our segmentation model, implemented within the MMSegmentation~\cite{mmseg2020} framework. Within our method (\autoref{fig:overview}) we always train the segmentation model from scratch to allow a fair comparison to other approaches. For further training details see our supplementary material or our codebase (released upon acceptance).

\subsection{Main Results}
In our main results, we compare our method of generating synthetic augmentations by simultaneously guiding the image generation process to produce more useful training samples with the corresponding non-guided approach introduced by \citet{kupyn2024dataset} at ECCV24. Additionally, we report results for the baseline model, which does not use any additional synthetic data. The methods are evaluated on two commonly used benchmark datasets: VOC2012 and COCO10k.
\begin{table*}[t]
\centering
\setlength{\tabcolsep}{8pt} 
\begin{tabular}{l l c c l l}
\toprule
Dataset & Method & Syn & Guided & mIoU & mAcc  \\
\midrule

\multirow{3}{*}{COCO10k}  
    & Baseline & \texttimes & \texttimes & 43.59 & 55.57 \\
    & ECCV24~\cite{kupyn2024dataset} & \checkmark & \texttimes & 44.05 {\small(+0.46)} & 55.88 {\small(+0.31)} \\
    & Ours & \checkmark & \checkmark & \textbf{45.44} {\small(+1.85)} & \textbf{56.91} {\small(+1.34)}  \\
\midrule
\multirow{3}{*}{VOC2012} 
    & Baseline & \texttimes & \texttimes & 91.42 & 95.07 \\
    & ECCV24~\cite{kupyn2024dataset} & \checkmark & \texttimes & 91.69 {\small(+0.27)} & 95.09 {\small(+0.02)} \\
    & Ours & \checkmark & \checkmark & \textbf{91.91} {\small(+0.49)} & \textbf{95.19} {\small(+0.12)} \\
\bottomrule
\end{tabular}

\caption{By incorporating guided synthetic data augmentation, our approach more than doubles the performance gain of \citet{kupyn2024dataset} on COCO10k, achieving an impressive +1.85 mIoU improvement over the baseline. Even on VOC2012, where performance is already near saturation, our method continues to push the boundaries, demonstrating measurable gains, again doubling the performance gain of~\citet{kupyn2024dataset}. These results highlight the effectiveness and generalizability of our approach in enhancing downstream model performance.}
\label{tab:main}
\end{table*}

Although~\citet{kupyn2024dataset} demonstrate promising improvements over the baseline, our method achieves twice the performance gain across all datasets, establishing a new state-of-the-art. The improvements on COCO10k are particularly substantial, showcasing the effectiveness of our approach. On VOC2012, where overall performance is already exceptionally high, further gains are inherently more challenging due to a possible performance ceiling. Nonetheless, our method still manages to achieve measurable improvements, highlighting its robustness even in near-saturated benchmarks.

\subsection{Ablations}
\begin{table*}[ht]
    \centering
    \begin{minipage}[b]{0.3\linewidth}
        \subcaptionbox{\label{tab:metrics_comparison}}{
        \centering
        \begin{tabular}{cccc}
        \toprule
         & MCD & CE & \textbf{Entropy} \\
        \midrule
        mIoU & 39.97 & 39.78 & \textbf{40.19} \\
        mAcc & 50.9 & 50.62 & \textbf{51.10} \\
        \bottomrule
        \end{tabular}
        }
    \end{minipage}
    \hfill
    \begin{minipage}[b]{0.3\linewidth}
        \subcaptionbox{\label{tab:pick_instance}}{
        \centering
        \begin{tabular}{ccc}
        \toprule
         & Most Certain & \textbf{Largest} \\
        \midrule
        mIoU & 39.45 & \textbf{40.19} \\
        mAcc & 50.93 & \textbf{51.10} \\
        \bottomrule
        \end{tabular}
        }
    \end{minipage}
    \hfill
    \begin{minipage}[b]{0.3\linewidth}
        \subcaptionbox{\label{tab:number_instance}}{
        \begin{tabular}{ccc}
        \toprule
        Object Count & 1 & \textbf{3} \\
        \midrule
        mIoU & 40.19 & \textbf{40.4} \\
        mAcc & 51.10 & \textbf{51.55} \\
        \bottomrule
        \end{tabular}
        }
    \end{minipage}
    \caption{Ablations of the active learning inspired ControlNet guidance by comparing downstream model performance when training with synthetically augmented data. (a) Comparison of different guidance loss functions (MCD, CE, Entropy), where entropy shows the most promising results. (b) Evaluation of augmenting the largest instance with guidance compared to augmenting the most certain instance. Augmenting the largest instance is a natural choice as it usually contributes more to the training loss. (c) Comparison of augmenting different numbers of objects with our method. Consistent with the ablation on object selection, augmenting larger regions of the image enhances performance.}
    \label{tab:ablations}
\end{table*}

\begin{figure*}[t]
  \centering
  \includegraphics[width=1.0\linewidth]{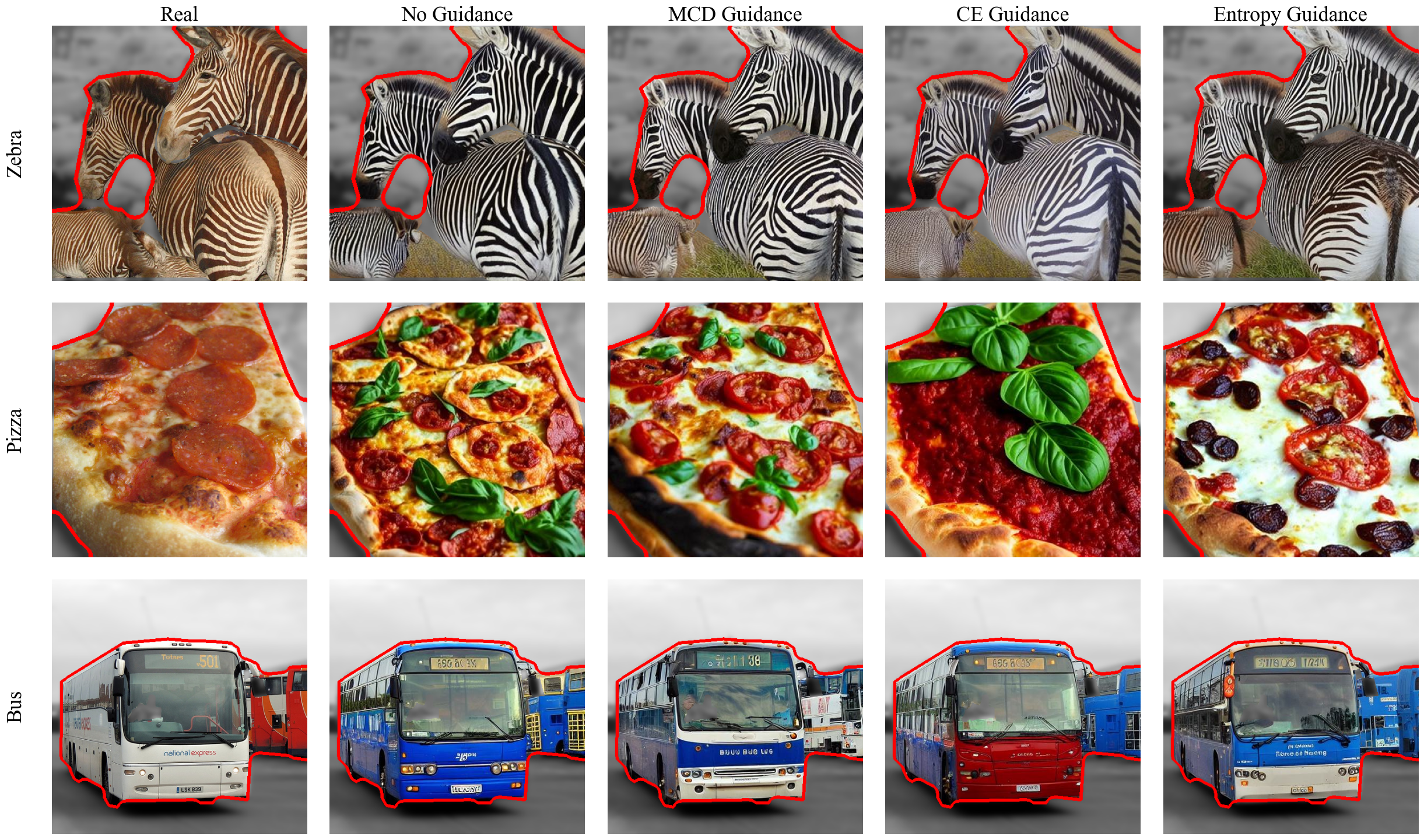}

   \caption{Qualitative comparison of different loss metrics during guidance. We visualize the real images (top row) as well as synthetically augmented images following~\cite{kupyn2024dataset} (second row) next to our proposed guidance in the following rows. Red borders outline the synthetically augmented object. The images share high visual quality.}
   \label{fig:metrics}
\end{figure*}

\begin{figure}[h]
  \centering
  \includegraphics[width=1.0\linewidth]{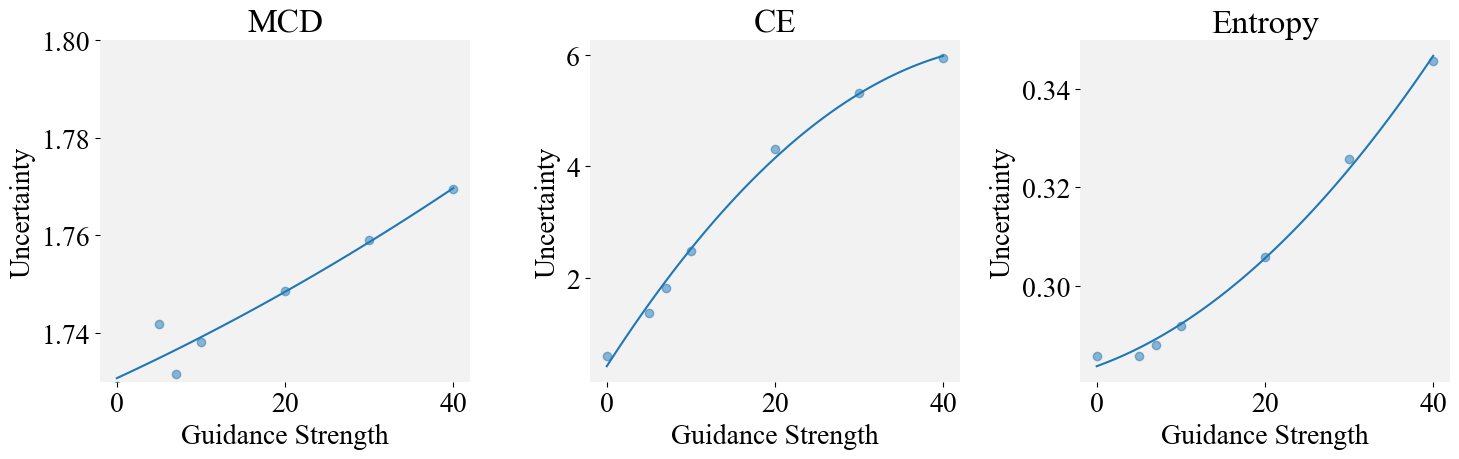}
   \caption{The plots visualize the uncertainty of generated objects when applying guided backward diffusion using different metrics at varying guidance strengths. Uncertainty is measured according to the respective metric. As guidance strength increases, we expect uncertainty to rise accordingly. However, MCD introduces noise, making it unreliable for gradient-based optimization. This instability arises from its stochastic nature, which injects variance into the optimization process, ultimately reducing its effectiveness.}
   \label{fig:mcd}
\end{figure}

\paragraph{Guidance Loss.}
Within this ablation, we investigate the suitability of different guidance metrics as defined in~\autoref{sec:method}. 
Despite the overall promising qualitative results in~\autoref{fig:metrics} we found one metric was outperforming the others: ~~\autoref{tab:metrics_comparison} shows the advantage of entropy. 
We attribute the advantage of entropy to two key limitations of the other metrics in the context of guidance. First, the randomness inherent in the MCD metric makes gradient-based optimization challenging, leading to noisy results, as illustrated in~\autoref{fig:mcd}. This figure visualizes the model's uncertainty on the generated data (y-axis) across different guidance strengths (x-axis) for each metric. Ideally, uncertainty should increase with stronger guidance, which holds true for both CE and entropy. However, MCD exhibits significant noise, making it unreliable for gradient-based optimization. This instability stems from the stochastic nature of MCD, which introduces variance into the optimization process, reducing its effectiveness.

\begin{figure}[h]
  \centering
  \includegraphics[width=1.0\linewidth]{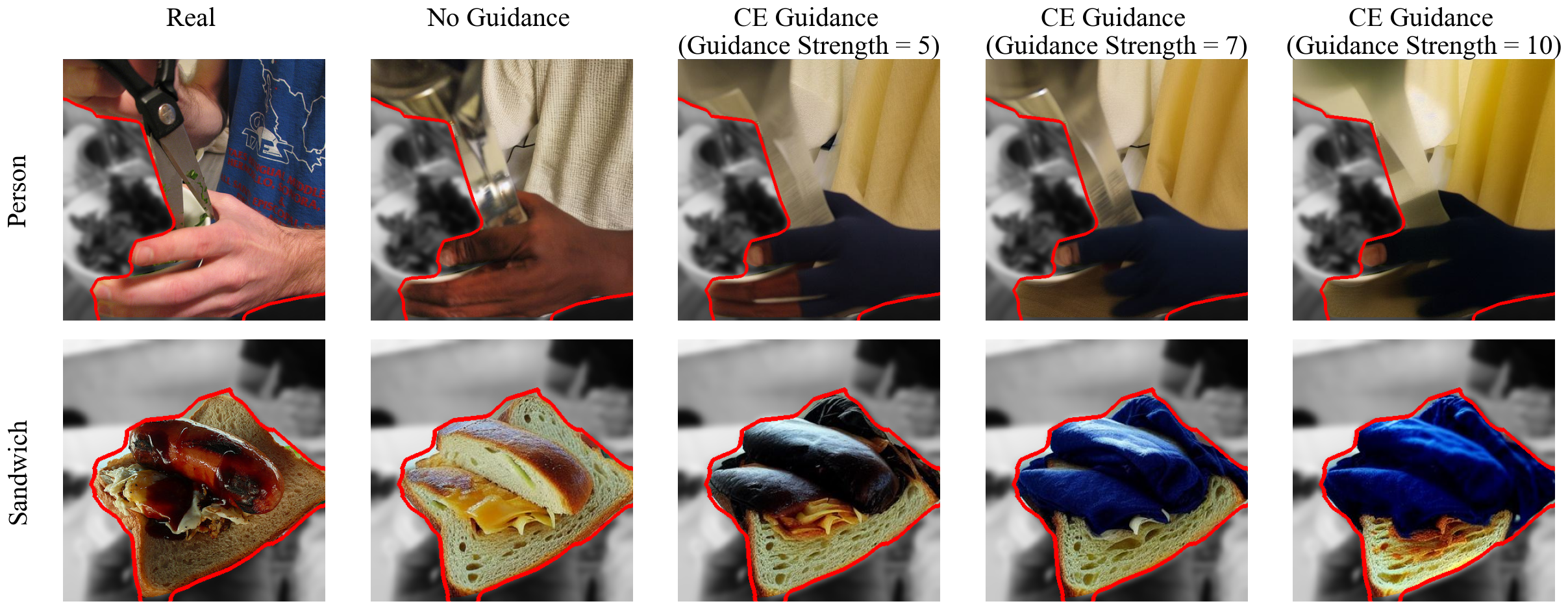}
   \caption{Visualization of failure cases of CE guidance. Guiding based on CE leads to an ill-posed optimization problem. It can encourage the prediction of images which are being misclassified by the model, but it might as well predict samples containing incorrect classes. }
   \label{fig:ce}
\end{figure}

Second, the CE loss formulation leads to an ill-posed optimization problem. While it encourages the generation of challenging samples—i.e., images where the model confidently predicts incorrect classes—it can also produce samples containing entirely incorrect classes, as demonstrated in~\autoref{fig:ce}. This unintended behavior reduces its effectiveness as a reliable guidance signal. 

\begin{figure*}[h]
  \centering
  \includegraphics[width=1.0\linewidth]{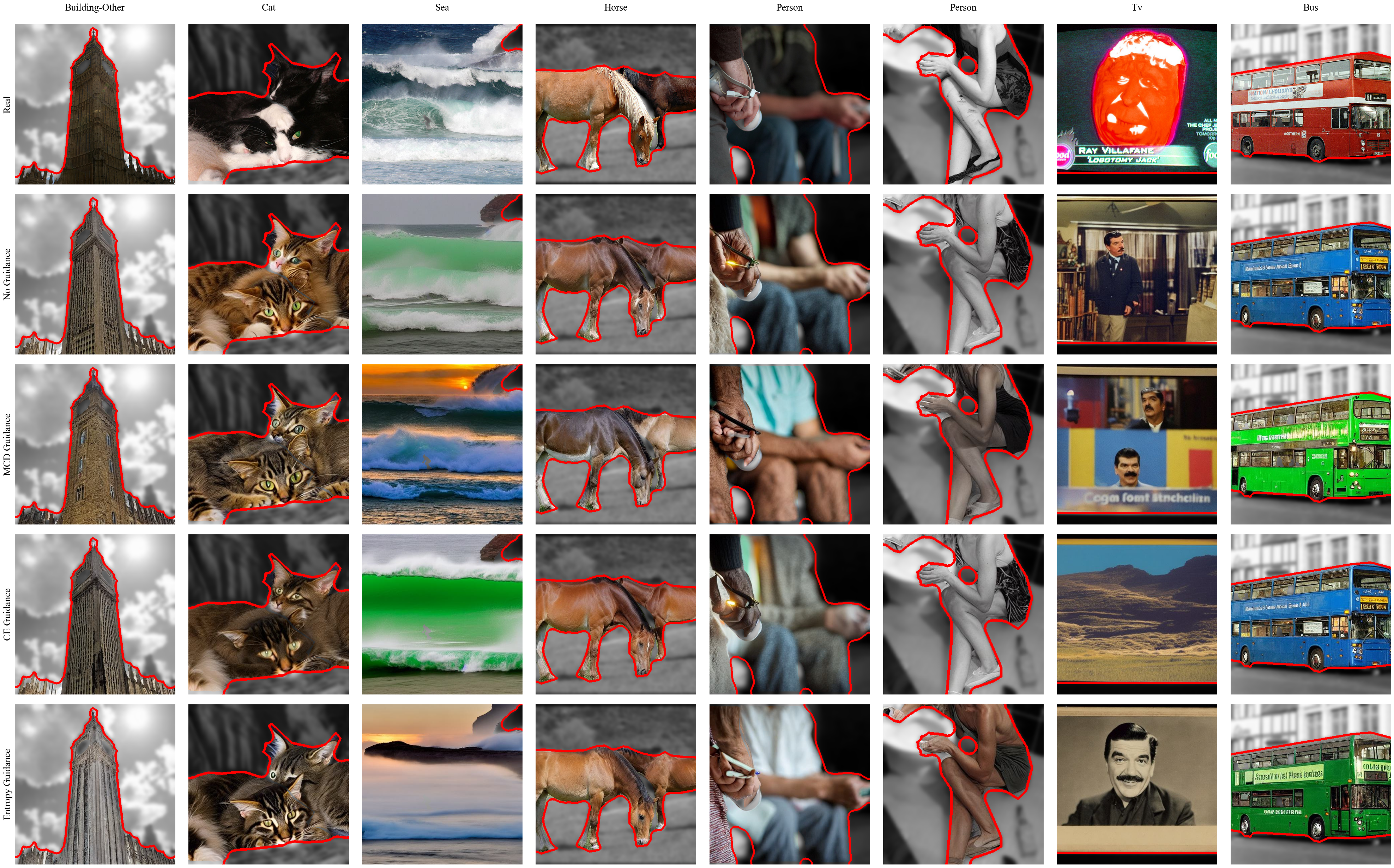}

   \caption{More qualitative results on the comparison of different loss metrics during guidance for multiple classes. We visualize the real images (top row) as well as synthetically augmented images following~\cite{kupyn2024dataset} (second row) next to our proposed guidance in the following rows. Red borders outline the synthetically augmented object. The images share high visual quality.}
   \label{fig:metrics1}
\end{figure*}

\begin{figure*}[h]
  \centering
  \includegraphics[width=1.0\linewidth]{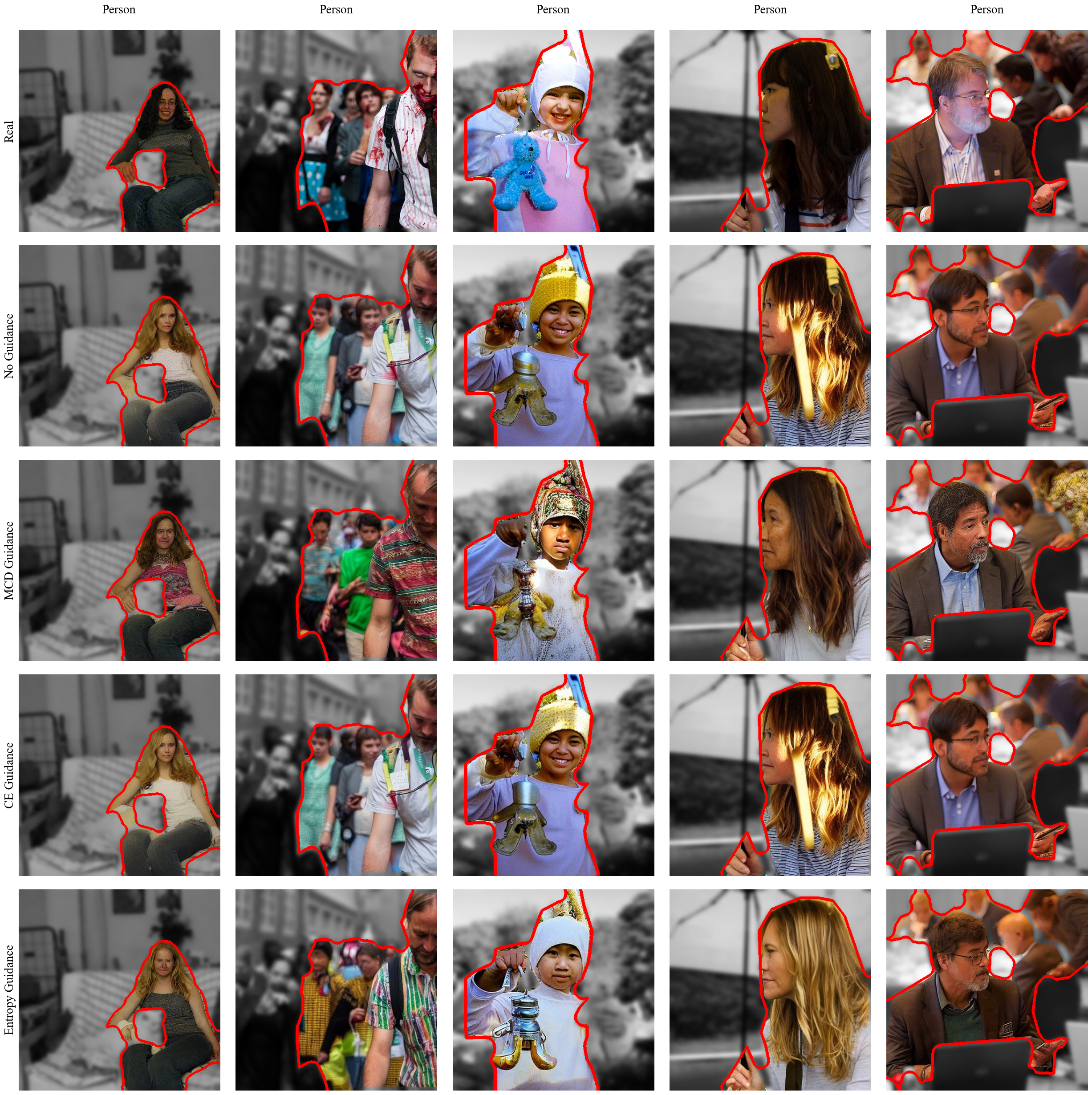}

   \caption{More qualitative results on the comparison of different loss metrics during guidance for class "person". We visualize the real images (top row) as well as synthetically augmented images following~\cite{kupyn2024dataset} (second row) next to our proposed guidance in the following rows. Red borders outline the synthetically augmented object. The images share high visual quality.}
   \label{fig:metrics2}
\end{figure*}

Hence, in all following experiments, we will leverage entropy guidance. For more qualitative evaluations please see our supplementary material at~\autoref{fig:metrics1} and~\autoref{fig:metrics2}.

\paragraph{Object Selection.}
As mentioned above,~\citet{kupyn2024dataset} propose redrawing each object individually and recombining them into the final image, ensuring perfect alignment with the existing ground truth mask. While highly effective, this approach is computationally expensive, especially for images containing many objects. To mitigate this, we explore alternative selection strategies that prioritize the most impactful objects for augmentation, aiming to maximize improvements in downstream model training. We focus on selection criteria based on object size and model uncertainty. Larger objects typically contribute more to the training loss, making them ideal for augmentation, so we prioritize selecting the largest objects. Additionally, inspired by active learning, we consider augmenting instances where the model is most certain. These objects only contribute little to the loss, and hence are expected to bring the greatest improvement when augmented, further enhancing the training process.

As shown in ~\autoref{tab:pick_instance}, selecting the largest instance remains more beneficial than selecting the most certain instance. This is likely because the most certain instances are often smaller, meaning their guided uncertainty would need to be significantly higher for them to have an equivalent impact than larger objects.

\paragraph{Number of Objects.}
The ablation study on instance selection suggests that augmenting larger objects leads to better model performance. To further explore this, we examine the effect of varying the number of augmented objects, which consequentially leads to larger augmented regions. As shown in ~\autoref{tab:number_instance}, our findings align with the previous experiment: augmenting larger regions of the image, hence augmenting more instances, improves performance.

\section{Conclusion}
In this work, we presented a novel paradigm for guiding the generation of synthetic training data introducing active learning inspired ControlNet guidance. By incorporating ControlNet guidance based on active learning criteria, we demonstrated that the value of synthetic data for downstream tasks can be substantially increased. Our results show that actively steering the diffusion process toward generating informative samples leads to superior segmentation model performance compared to unguided data generation. This shifts the conventional approach from post hoc sample selection to proactive, task-aware data synthesis, fundamentally redefining how synthetic datasets are constructed.

Our results highlight that entropy-based guidance outperforms other criteria, emphasizing the importance of uncertainty-based selection in generative modeling. However, our method introduces computational overhead due to segmentation model training and guidance, motivating research into efficient surrogates. Additionally, the strength of the guidance is a crucial hyperparameter—if too weak, the effect is negligible; if too strong, the generated images may deviate from the original distribution. While this could impact immediate model performance, it may also enhance generalization, which remains an avenue for future work. 

Beyond these considerations, our work lays the foundation for a broader research direction. Future studies could, again inspired by active learning, investigate iterative, cyclic approaches where the segmentation model continuously refines itself through newly generated data, akin to curriculum learning. This could unlock further improvements in model robustness and efficiency.

Overall, this work presents a transformative approach to dataset generation, demonstrating that generative models can do more than merely mimic real-world data—they can be harnessed to produce training samples that are intrinsically more valuable. As generative models continue to improve, this perspective will be critical in shaping the future of data-driven deep learning, reducing reliance on expensive manual annotations, and enabling more efficient AI systems.

{
    \small
    \bibliographystyle{ieeenat_fullname}
    \bibliography{main}
}

\clearpage
\setcounter{page}{1}

\twocolumn[{%
\renewcommand\twocolumn[1][]{#1}%
\maketitlesupplementary
\captionsetup{type=figure}
\begin{center}
        \begin{minipage}{\textwidth}
            \centering
            \subcaptionbox{\label{fig:scheduler}}{%
                \includegraphics[width=0.9\linewidth]{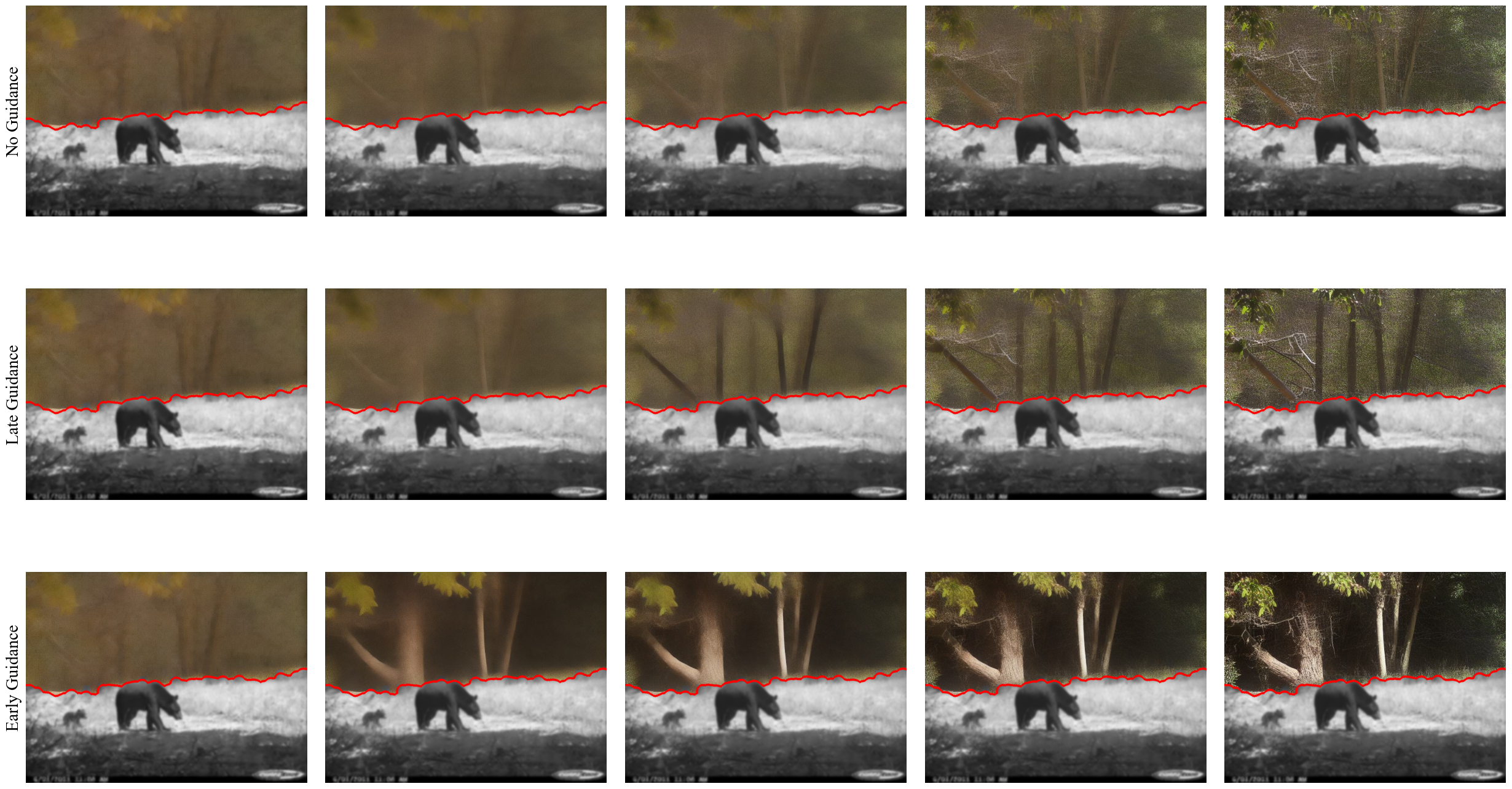}
            }
        \end{minipage}
        \vspace{0.5cm} 
    
        
        \begin{minipage}{0.29\textwidth}
            \centering
            \subcaptionbox{\label{fig:scheduler-plot}}{%
                \includegraphics[width=\linewidth]{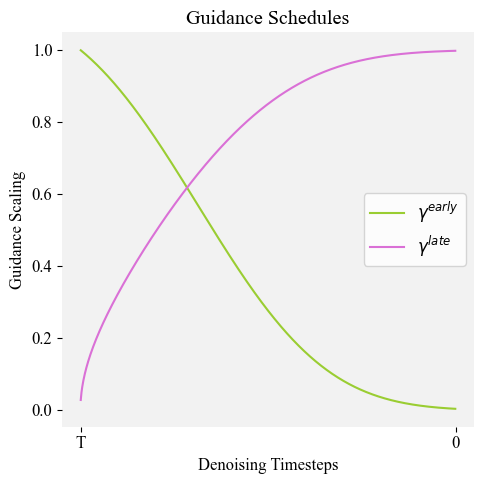}
            }
        \end{minipage}
        \begin{minipage}{0.29\textwidth}
            \centering
            \subcaptionbox{\label{fig:scheduler-results1}}{%
                \includegraphics[width=\linewidth]{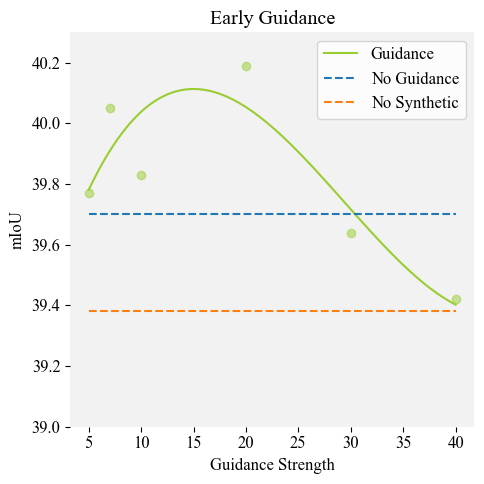}
            }
        \end{minipage}
        \begin{minipage}{0.29\textwidth}
            \centering
            \subcaptionbox{\label{fig:scheduler-results2}}{%
                \includegraphics[width=\linewidth]{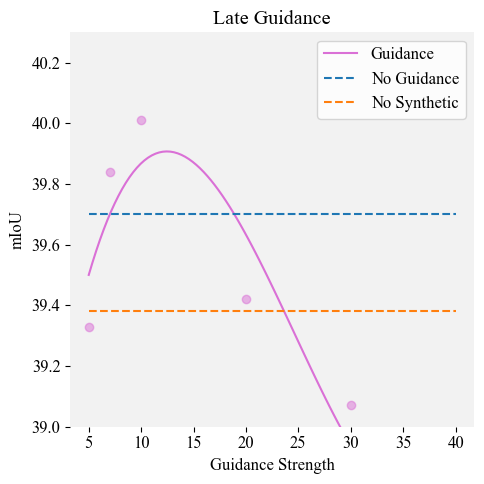}
            }
        \end{minipage}
        \end{center}%

        \caption{
        (a) Qualitative assessment of different guidance schedules. The figure shows the approximated $\hat{x_0}$ at different denoising iterations from left to right. The top row depicts unguided backward diffusion, while the middle and bottom rows illustrate guided backward diffusion using $\gamma^{late}$ and $\gamma^{early}$, respectively. Applying $\gamma^{early}$ can lead to structural modifications in object features, whereas $\gamma^{late}$ primarily affects texture.
        (b) Visualization of the early guidance ($\gamma^{early}$) and the late guidance ($\gamma^{late}$) schedulers to control the guidance strength for each denoising step individually. (c) Downstream model performance when trained on data generated with $\gamma^{early}$ at different guidance strengths. The orange line represents performance without synthetic augmentation, while the blue line corresponds to performance with synthetic augmentation following~\cite{kupyn2024dataset}. (d) Same as (c), but using $\gamma^{late}$. 
        These results suggest that model training benefits more from structural modifications introduced by $\gamma^{early}$ rather than textural changes. Textural modifications tend to produce unnatural images more quickly, even when imperceptible to the human eye—similar to JPEG compression, which removes high-frequency details due to their limited visibility, or adversarial perturbations, which introduce imperceptible high-frequency changes to manipulate the image—ultimately leading to quicker performance degradation.
        }

}]

\twocolumn[{%
    \section{Guidance Schedule}\label{supp:schedule}
    Guiding the backward diffusion process at different denoising timesteps can yield varying effects on the generated image. We therefore investigate the impact of applying stronger guidance during early iterations compared to later iterations. We base the schedulers on the noise schedule of the forward diffusion process. We define $\gamma^{early}(t) = \alpha_t$ and $\gamma^{late}(t) = \sqrt{1-\alpha_t}$, where $\alpha_t$ represents the cumulative product of the noise schedule's variance values at time step $t$. For a visualization of the schedules, see~\autoref{fig:scheduler-plot}.

    As shown in~\autoref{fig:scheduler}, we observe that $\gamma^{early}$ significantly alters the structural features of the objects, while $\gamma^{late}$ primarily affects texture.

    In our quantitative experiment, we compare no synthetic data augmentation to non-guided augmentations based on Kupyn et al.'s work~\cite{kupyn2024dataset}, and our proposed guided data generation using the entropy loss function. We test different guidance scales with the introduced schedulers and present the results in~\autoref{fig:scheduler-results1} and~\autoref{fig:scheduler-results2}. Although the images generated with both schedules appear to have high quality, as shown in~\autoref{fig:scheduler}, we found that early guidance, $\gamma^{early}$, has a more pronounced impact than late guidance, $\gamma^{late}$ (see~\autoref{fig:scheduler-results1} and~\autoref{fig:scheduler-results2}). Furthermore, we observed that the overall guidance strength can be higher for early guidance, which highlights the effectiveness of $\gamma^{early}$ in generating more natural changes in the image content, preventing overly aggressive, unnatural shifts in the generated images. Notably, when the guidance strength is too high, the generated images deviate from the training/validation data distribution, resulting in performance degradation.

    \section{Implementation Details}
    
    \paragraph{Segmentation Model.} For training the segmentation model, we adopt the AdamW optimizer with an initial learning rate of $6 \times 10^{-5}$, following a poly learning rate (PolyLR) schedule. We use a batch size of 16. The maximum training iteration is dataset-dependent:  

    \begin{itemize}
        \item COCO2.5k: 40,000 iterations  
        \item COCO10k: 80,000 iterations  
        \item VOC2012: 40,000 iterations
    \end{itemize}

    \paragraph{Data Generation.}
    For data generation with ControlNet, we utilize a pretrained model from~\citet{kupyn2024dataset} using the Stable Diffusion backbone. Similar to~\citet{kupyn2024dataset} we incorporate $T=40$ diffusion denoising steps. The guidance is conditioned on predicted depth and HED edge detection, ensuring structure-aware synthesis. Our generation pipeline employs an image size of $768$. To balance fidelity and control strength, we vary the guidance scale between 6.0 and 7.5, while the conditioning scale is set at 0.7 and 0.2. As the guidance strength is an important parameter in our active learning inspired ControlNet guidance, we test different guidance strength for the different datasets, where $\eta \in [7,10,20,30]$. We finally used following setup: 
    \begin{itemize}
        \item COCO2.5k: $\eta = 20$
        \item COCO10k: $\eta = 7$ 
        \item VOC2012: $\eta = 10$ 
    \end{itemize}

    For further details, please see our code base (released upon acceptance).  

}]

\onecolumn
\newpage
\section{Visualizations}

\end{document}